\documentclass[letterpaper, 10 pt, journal, twoside]{ieeetran}
\usepackage{amsmath,amsfonts}
\usepackage{algorithmic}
\usepackage{array}
\usepackage[caption=false,font=normalsize,labelfont=sf,textfont=sf]{subfig}
\usepackage{textcomp}
\usepackage{stfloats}
\usepackage{url}
\usepackage{verbatim}
\usepackage{graphicx}
\usepackage{xcolor}
\usepackage{hyperref}    
\usepackage{float}
\usepackage{subcaption}
\usepackage{gensymb}
\usepackage{amssymb}
\usepackage{enumitem}
\def\BibTeX{{\rm B\kern-.05em{\sc i\kern-.025em b}\kern-.08em
    T\kern-.1667em\lower.7ex\hbox{E}\kern-.125emX}}
\usepackage{balance}
\usepackage[
  backend=biber,
  style=ieee,     
  sorting=none 
]{biblatex}
\begin{document}
\title{
Learning Category-level Last-meter Navigation from RGB Demonstrations of a Single-instance}

\author{Tzu-Hsien Lee, Fidan Mahmudova and Karthik Desingh \\University of Minnesota, Twin Cities \vspace{-0.5cm}
}

\maketitle

\begin{abstract}
Achieving precise positioning of the mobile manipulator's base is essential for successful manipulation actions that follow. Most of the RGB-based navigation systems only guarantee coarse, meter-level accuracy, making them less suitable for the precise positioning phase of mobile manipulation. 
This gap prevents manipulation policies from operating within the distribution of their training demonstrations, resulting in frequent execution failures. We address this gap by introducing an object-centric imitation learning framework for \textit{last-meter navigation}, enabling a quadruped mobile manipulator robot to achieve manipulation-ready positioning using only RGB observations from its onboard cameras. Our method conditions the navigation policy on three inputs: goal images, multi-view RGB observations from the onboard cameras, and a text prompt specifying the target object. A language-driven segmentation module and a spatial score-matrix decoder then supply explicit object grounding and relative pose reasoning. Using real-world data from a single object instance within a category, the system generalizes to unseen object instances across diverse environments with challenging lighting and background conditions. To comprehensively evaluate this, we introduce two metrics: an edge-alignment metric, which uses ground truth orientation, and an object-alignment metric, which evaluates how well the robot visually faces the target. Under these metrics, our policy achieves average success rates of 74.58\% in edge-alignment and 89.42\% in object-alignment when positioning relative to unseen target objects. These results show that precise last-meter navigation can be achieved at a category-level without depth, LiDAR, or map priors, enabling a scalable pathway toward unified mobile manipulation. Further details, visualizations, and videos are provided on the project page at \url{https://rpm-lab-umn.github.io/category-level-last-meter-nav/}
\end{abstract}


\section{INTRODUCTION}

\IEEEPARstart{F}{or} robots to integrate effectively into human environments and perform useful tasks, they must be capable of both moving through the space and interacting with surrounding objects. This integration of navigation and manipulation, commonly referred to as mobile manipulation, is a fundamental capability for autonomous assistive robots. Crucially, manipulation policies operate reliably only when the robot’s position and orientation fall within the distribution covered by their demonstration data~\cite{yang2025mobipi}. Existing navigation approaches, however, are designed around much coarser requirements: most global navigation benchmarks define success as stopping within roughly 1 meter of the target location~\cite{zeng2024poliformer, chang2023goat, yokoyama2024vlfm, xue2025omninav, rajvanshi2024saynav, gervet2023navigating, habitatchallenge2023}. Even the widely adopted navigation evaluation protocol from Anderson et al.~\cite{anderson2018evaluation} similarly specifies success as reaching within twice the agent’s body width, typically about 1 meter from the target location. As a result, navigation systems often fail to position the robot precisely enough for the manipulation system to follow and succeed, leading to frequent failures during task execution~\cite{ehsani2024spoc}. This mismatch reveals a critical gap between these phases of mobile manipulation.

\begin{figure}[!t]
    \centering
    \includegraphics[width=\linewidth]{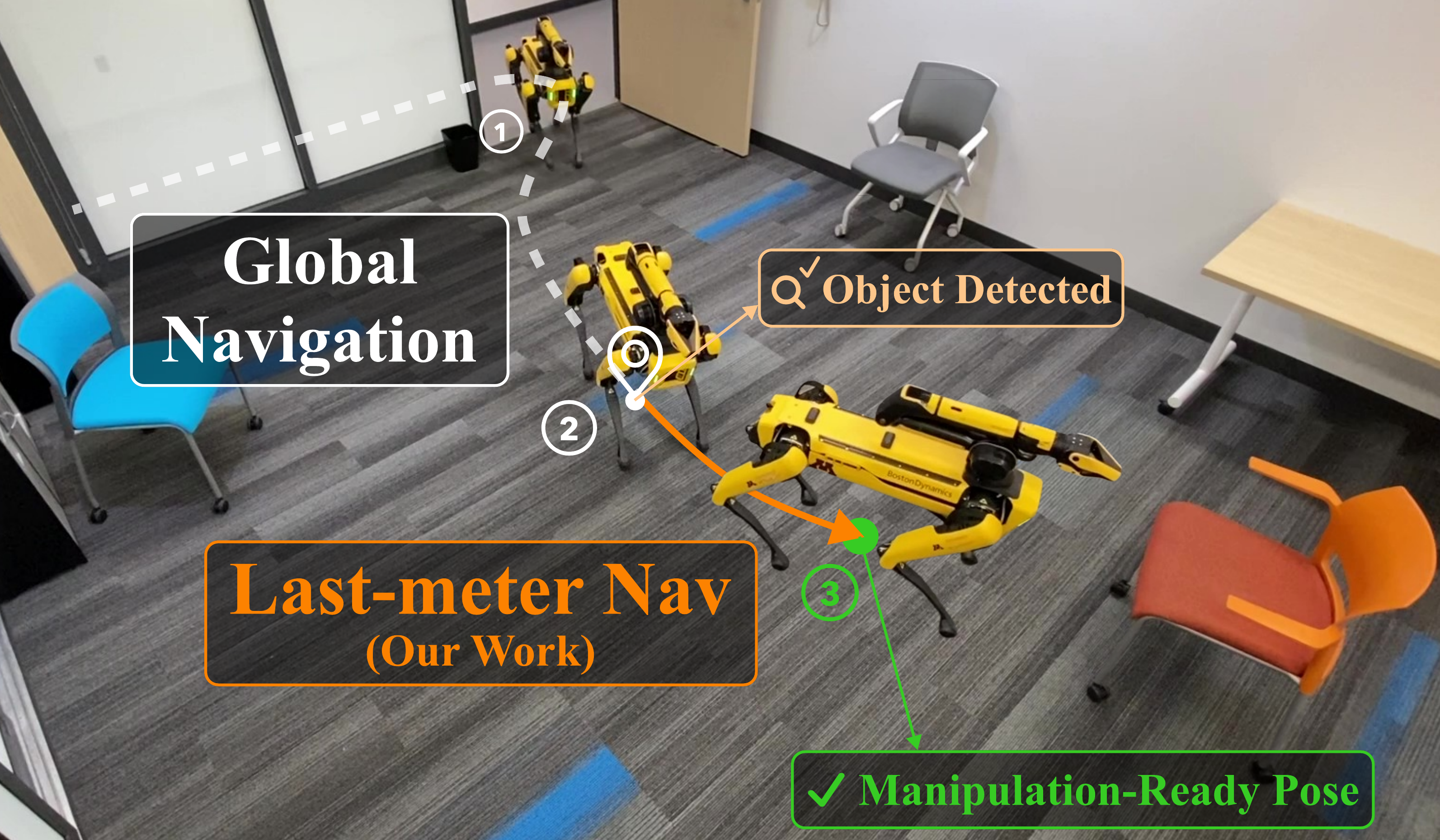}
    \caption{\footnotesize{\textbf{Last-meter navigation bridges the gap between global navigation and manipulation.} We propose an object-centric imitation learning framework using onboard multi-view RGB observations to enable last-meter navigation. \textit{In this example}, (1) Global navigation brings the robot to the vicinity of the target. (2) Upon detecting the target object (e.g., the orange chair), our policy is invoked. (3) \textbf{Last-meter navigation} refines the robot's position and orientation to a precise manipulation-ready pose defined by a goal observation, remaining robust in the presence of distractors.}}
    \label{fig:teaser_figure}
    \vspace{-0.5cm}
\end{figure}

To achieve centimeter-level positioning precision, navigation systems often rely on strong prior knowledge of the environment, such as high-resolution maps or neural radiance fields (NeRFs)~\cite{wang2023nerf, adamkiewicz2022vision, xue2025omninav}, or require additional sensing modalities such as depth cameras, LiDAR, or odometry~\cite{chang2023goat, yokoyama2024vlfm, rajvanshi2024saynav, gervet2023navigating}. These requirements for prior knowledge and additional sensing modalities limit scalability in real-world deployment: environmental priors are costly to obtain and often brittle in dynamic or movable-target settings, while extra sensors add hardware cost and system complexity. In contrast, RGB cameras are inexpensive, already standard on most commercial and research robots, and often sufficient for many perception tasks, making RGB-only policies an appealing direction.

Recent navigation research therefore emphasizes RGB-based, end-to-end learning methods, either via imitation learning or reinforcement learning~\cite{ehsani2024spoc, zeng2024poliformer}, and vision-language foundation models that leverage large-scale pretraining for semantic navigation~\cite{yokoyama2024vlfm}. However, these methods still adopt coarse success metrics (e.g., one-meter thresholds), leaving their precision inadequate for downstream manipulation tasks that demand centimeter-level accuracy.

To close this gap, we define the ``last-meter navigation'' problem, a focused stage between global path planning and manipulation execution, whose objective is to achieve centimeter-level precise position and degree-level orientation alignment relative to the target location. To this end, we adopt two simplifying assumptions that isolate the last-meter navigation problem from broader navigation challenges such as exploration or obstacle avoidance:
1) The target object remains within the robot’s field of view throughout the last-meter navigation.
2) This navigation is free of occluding obstacles.

Designing last-meter navigation with hand crafted rules is difficult to scale. For example, specifying fixed thresholds for how large the object should appear in the image, how centered it must be, or how the robot should rotate when the object shifts in the camera view quickly breaks down across different object shapes, surface textures, lighting conditions, backgrounds, and camera perspectives. Human environments are visually diverse and often ambiguous, and without depth or reliable metric information, manually encoding how the robot should interpret these visual cues and convert them into precise motions becomes brittle and unreliable.

Therefore, we propose an object-centric, data-driven approach that enables the robot to learn spatial relationships and control behaviors directly from raw RGB observations. In particular, we employ imitation learning, leveraging expert demonstrations to efficiently learn goal-directed navigation policies. Imitation learning has demonstrated strong performance in robotic manipulation, where it captures fine-grained, object-centric behaviors from demonstrations. Extending this successful manipulation learning framework to last-meter navigation establishes a unified foundation for mobile manipulation, allowing the same policy family to govern both movement toward and physical interaction with target objects.

A key bottleneck in imitation learning is the high cost of collecting demonstration data, especially in real-world environments. To address this, our method is trained on a single instance of an object category and is designed to generalize to unseen instances within the same category. We use chairs as a representative example due to their abundance and visual diversity.

Empirically, our method attains a 94\% success rate on the seen instance and 74.58\% on unseen instances under ground-truth evaluation with strict translation and orientation thresholds. Under visual evaluation metrics, performance increases to 96.67\% on the seen instance and 89.42\% on unseen instances. Together, these results demonstrate strong instance-to-category generalization, achieving manipulation-ready precision across diverse object appearances using RGB input alone.

The contributions of this paper include:
\begin{enumerate}[leftmargin=*]
    \item An object-centric imitation learning framework that solves the last-meter navigation problem and produces manipulation-ready base poses.
    \item A demonstration of strong instance-to-category generalization, where a model trained on a single object instance reliably transfers to unseen objects of the same category.
    \item A real-world validation that precise last-meter navigation is achievable using only onboard RGB observations, without depth, LiDAR, or map priors. This design choice reflects practical scenarios where additional sensor modalities may be unavailable, unreliable, or cost-prohibitive.
\end{enumerate}
\section{Related Work}
Recent studies in robot navigation and manipulation have advanced both capabilities, yet precise coordination between them remains underexplored. To contextualize our work, we briefly review three relevant directions.

\subsection{RGB-only Learning-based Global Navigation}
Recent end-to-end learning approaches have enabled robots to navigate complex and visually diverse environments using only RGB observations. PoliFormer~\cite{zeng2024poliformer} presents a transformer-based on-policy reinforcement learning framework that scales efficiently to long-horizon navigation and achieves state-of-the-art results in challenging visual scenes. In a related approach, NoMaD~\cite{sridhar2024nomad} employs goal-conditioned diffusion policies that jointly model exploration and navigation, producing robust trajectories without explicit maps or depth input. While many other navigation systems leverage additional sensing modalities such as depth, LiDAR, or pre-built maps to simplify localization and planning~\cite{chang2023goat, tang2025openin, wang2023nerf, xue2025omninav, adamkiewicz2022vision, rajvanshi2024saynav, yokoyama2024vlfm, gervet2023navigating}, PoliFormer and NoMaD demonstrate that RGB-only visual inputs combined with high-capacity learning architectures can still produce reliable global navigation policies. However, they typically focus on reaching the general vicinity of the goal, rather than achieving the fine-grained alignment necessary for manipulation. Our work addresses this limitation by explicitly focusing on the last-meter phase, where centimeter-level precision is essential for interaction with target objects.

\subsection{Positioning Robot Base for Mobile Manipulation}
Accurate robot base positioning is essential for reliable manipulation performance. Mobi-$\pi$~\cite{yang2025mobipi} clearly demonstrates that manipulation policies fail when the robot’s base pose falls outside the distribution of the training data, underscoring the importance of precise positioning. Similarly, optimization-based studies such as MoMa-Pos~\cite{shao2024moma} and ~\cite{zhang2023base} consistently emphasize that the choice of base location directly affects reachability, manipulability, and overall task success. Together, these findings confirm that base placement is a key determinant of manipulation success in mobile manipulation systems. Integrated frameworks such as SPOC~\cite{ehsani2024spoc} further highlight this dependency: despite combining navigation and manipulation within a unified policy, their reported failures often stem from slight misalignments between the gripper and the target object, reinforcing the critical need for accurate last-meter navigation.

\subsection{Last-meter Navigation}
Several recent works have begun addressing the last-meter navigation stage that directly precedes manipulation. Aim My Robot~\cite{meng2025aim} introduces a precision local navigation policy that guides the robot toward arbitrary target objects with centimeter-level accuracy, using depth and LiDAR sensing for geometric reasoning. MoTo~\cite{wu2025moto} extends this direction with an interaction-aware navigation policy that generates feasible docking points via object-centric reasoning. However, MoTo relies on depth sensing and an offline pre-scanning phase to construct a global scene graph, which prevents online adaptation in dynamic environments. While these methods successfully target the last-meter stage, their reliance on heavy sensing modalities and static global maps restricts general deployment. In contrast, our approach attains comparable precision using only RGB input, enabling broader applicability across diverse robots and settings.

\section{Methodology}

\begin{figure*}[t]
    \centering
    \includegraphics[width=\textwidth]{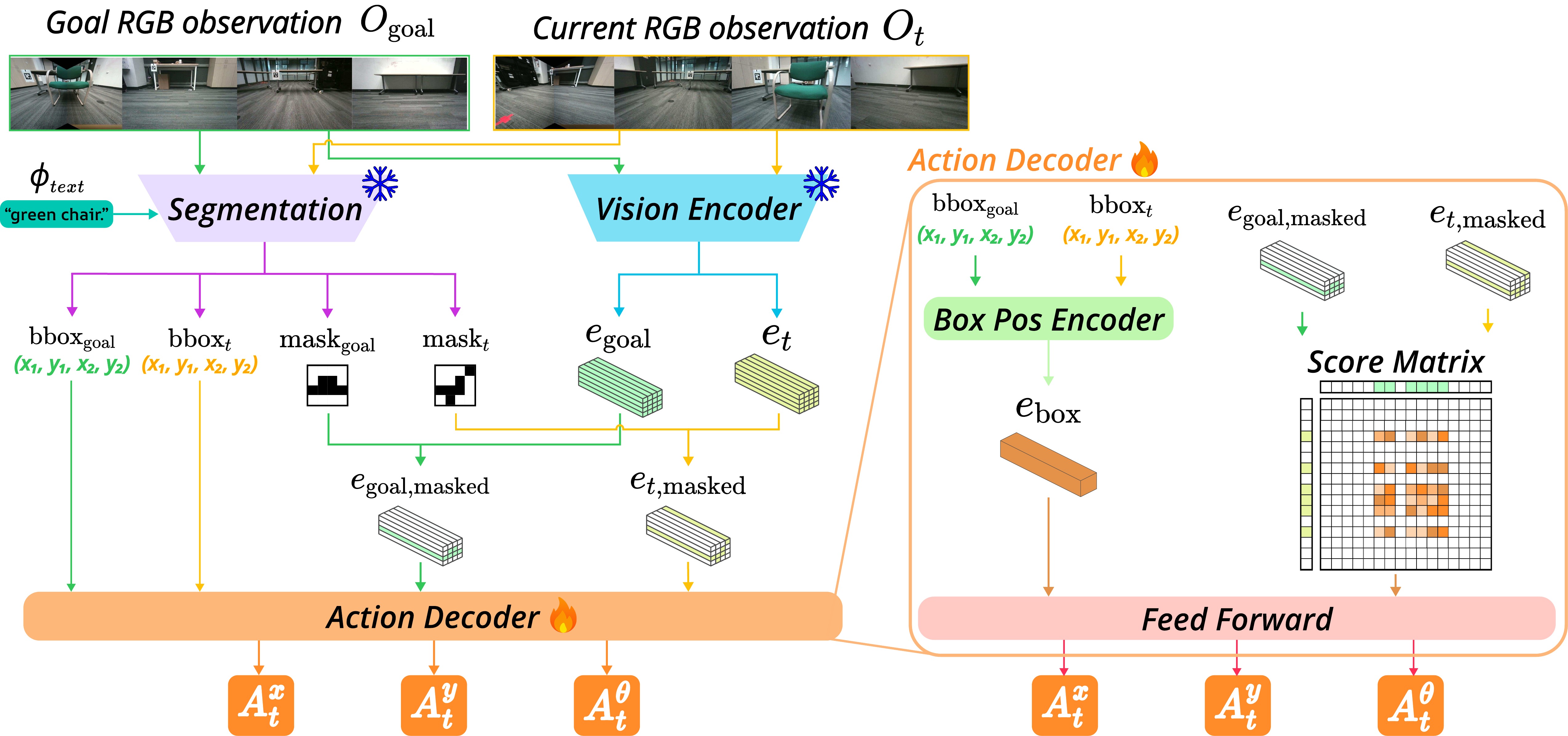}
    \caption{\footnotesize{\textbf{Architecture Overview.} At each timestep $t$, the model receives the current observation $O_t$ and the goal observation $O_{\text{goal}}$ and encodes them into feature embeddings $e_t$ and $e_{\text{goal}}$. The segmentation module, conditioned on the language prompt $\phi_{\text{text}}$, produces object masks and bounding boxes. The masked and cropped embeddings ($e_{t,\text{masked}}$, $e_{\text{goal},\text{masked}}$), together with the bounding box coordinates, are passed to the action decoder. Inside the decoder, the box coordinates are encoded into a box embedding $e_{\text{box}}$, while $e_{t,\text{masked}}$ and $e_{\text{goal},\text{masked}}$ are used to compute the similarity score matrix. The box embedding and the flattened score matrix are then concatenated and fed into a feedforward network to output the predicted action $A_t$.}}
    \label{fig:architecture}
\end{figure*}

\subsection{Problem Formulation}
We formulate the last-meter navigation problem as a behavioral cloning (BC) task, where a policy $\pi$ is trained to imitate expert demonstrations. Specifically, the policy maps multi-modal inputs to an action at time $t$:

$$\pi: (O_t,\;O_{goal}, \;\phi_{\text{text}})\to{A_t}$$

where $O_t$ represents the current multi-view RGB observations, $O_{goal}$ denotes the goal multi-view RGB observations captured at the desired terminal pose, $\phi_{\text{text}}$ is a natural-language prompt specifying the target object, and $A_t$ is the current discrete action selected by the policy.

Each RGB observation consists of images from the robot’s onboard cameras, covering a 360-degree view:
$$O \in \{I^{front}, I^{right}, I^{back}, I^{left}\}$$

The action $A_t$ consists of three motion primitives: forward ($x$), lateral ($y$), and rotational ($\theta$), represented as:~$$A_t \in \{A_t^x, A_t^y, A_t^{\theta}\}$$

Each primitive is discretized into three levels: negative, zero, and positive. Negative corresponds to motion in the negative direction, zero corresponds to no motion, and positive corresponds to motion in the positive direction. The step sizes are set to $0.2$m for translation and $12\degree$ for orientation, representing the minimum commanded increments that the Spot robot can reliably execute. These step sizes therefore determine the precision floor, which is reflected in the data collection and evaluation tolerances reported in Section~\ref{sec:data_collection}.

Following the BC paradigm, the policy parameters $\omega$ are optimized by minimizing the negative log-likelihood of expert actions over the training dataset ($D$):

$$\mathcal{L}(\omega) = -\mathbb{E}_{(O_t, O_{\text{goal}}, \phi_{\text{text}}, A_t^*) \sim {D}} \bigl[\;\log \pi_\omega\bigl(A_t^* \mid O_t, O_{\text{goal}}, \phi_{\text{text}}\bigr)\bigr]$$

where $A_t^*$ is the expert action at time $t$ (see Section~\ref{sec:data_collection}).

\subsection{Goal Conditioning}
We condition the policy on both goal observations $O_{\text{goal}}$ and a text prompt $\phi_{\text{text}}$. The text prompt specifies the target object, enabling the model to focus on the relevant object in the observations. The goal observations provide visual information about the desired final pose, allowing the policy to infer the goal position and orientation relative to the target object. 

The assumption regarding the availability of goal observations holds in many practical scenarios. For navigation tasks, such observations can be captured during the mapping phase; for mobile manipulation tasks, they can be drawn directly from the training dataset of the manipulation policy, where the goal pose is already specified or demonstrated.

\subsection{Architecture}
\label{sec:architecture}
Recent advances in vision foundation models provide powerful pretrained encoders that map images into semantically rich embeddings~\cite{oquab2024dinov, darcet2023vitneedreg, radford2021learning, he2022masked}. Our system leverages these pretrained representations to reduce the amount of task-specific training data, especially when there is no simulation environment available. The proposed architecture consists of three modules: (1) a vision encoder, (2) a segmentation module, and (3) an action decoder (Fig.~\ref{fig:architecture}).

Current and goal observation images are first processed by the vision encoder to obtain feature embeddings. In parallel, the segmentation module takes the language prompt along with both current and goal observations and generates object masks with corresponding bounding boxes. Using this information, the embeddings are cropped to the bounding box region and masked to retain only the target object embeddings. The resulting masked embeddings, together with the bounding box coordinates, are provided to the action decoder, which predicts the next-step action.

\subsubsection{\textbf{Vision Encoder}}
We adopt DINOv2~\cite{oquab2024dinov, darcet2023vitneedreg} as the vision encoder to extract robust, semantically aware image features from both the current and goal observations.

Each image $\mathcal{I}^{640 \times 480}$ from $O_t$ and $O_{\text{goal}}$ is processed individually, producing embeddings $e_t$ and $e_{\text{goal}}$, each in $\mathbb{R}^{34 \times 34 \times 1024}$.

The encoder is used in a frozen, pretrained form without finetuning, ensuring that its general-purpose semantic representations are preserved. This choice reduces computational cost, prevents overfitting to the relatively small navigation dataset, and maintains the strong generalization properties of large-scale pretraining, including robustness across object categories.

\subsubsection{\textbf{Segmentation Module}}
To highlight the target object, we employ a two-stage text-driven segmentation process. OwlV2~\cite{minderer2023scaling} detects the object specified by the language prompt and produces a bounding box, which is then refined by SAM2~\cite{ravi2024sam} to generate a segmentation mask. The resulting bounding box and mask are passed downstream to crop and filter the visual embeddings for action prediction.

\subsubsection{\textbf{Action Decoder}}
The decoder integrates bounding box information with masked embeddings to produce discrete action predictions $A_t$. Bounding box coordinates are concatenated and projected into a box embedding $e_{\text{box}}$ of dimension $\mathbb{R}^{4096}$ using a multi-layer perceptron(MLP). Masked embeddings $e_{t,\text{masked}}$ and $e_{\text{goal,masked}}$ are spatially pooled to $8 \times 8$ feature maps, yielding 64 spatial tokens per observation. The score matrix $S \in \mathbb{R}^{64 \times 64}$ is computed as the pairwise cosine similarity between current and goal tokens, where $S_{i,j}$ encodes the visual correspondence between location $i$ in the current view and location $j$ in the goal view. The matrix therefore forms a dense correspondence map whose pattern of high-similarity entries reflects how the current view is spatially offset from the goal. The $8 \times 8$ pooling resolution was chosen for memory efficiency; finer resolutions (e.g., $16 \times 16$) would yield a $256 \times 256$ matrix at substantially higher memory and computational cost. The flattened score matrix is then concatenated with the box embedding and passed through an MLP that outputs three 3-class classification heads, corresponding to three motion primitives: $A_t^x$, $A_t^y$, $A_t^{\theta}$.

\subsection{Auxiliary Stopping Mechanism}
\label{sec:stopping_mechanism}

A rollout terminates when the policy predicts two consecutive stop actions $A_t = \{0, 0, 0\}$. To address the precise-termination issue discussed in Section~\ref{sec:policy_comparison}, we introduce a rule-based auxiliary stopping mechanism that overrides $A_t$ with the stop action when visual alignment between the current and goal observations is satisfied.

At each timestep $t$, before $A_t$ is executed, we compute two signals from the segmentation module: the IoU between the current and goal masks $\mathcal{M}_t, \mathcal{M}_{\text{goal}}$, and the absolute difference in their centers of mass along $x$, $|c_t^x - c_{\text{goal}}^x|$. The stop is triggered when $\mathrm{IoU}(\mathcal{M}_t, \mathcal{M}_{\text{goal}}) > 0.6$ and $|c_t^x - c_{\text{goal}}^x| < 25$ pixels both hold; otherwise $A_t$ is executed unchanged. The mechanism is a deterministic inference-time override and does not affect training. The same thresholds are used across all evaluated categories.
\section{Experiment Setup}

\subsection{Data Collection}
\label{sec:data_collection}
\begin{figure}
    \centering
    \includegraphics[width=\linewidth]{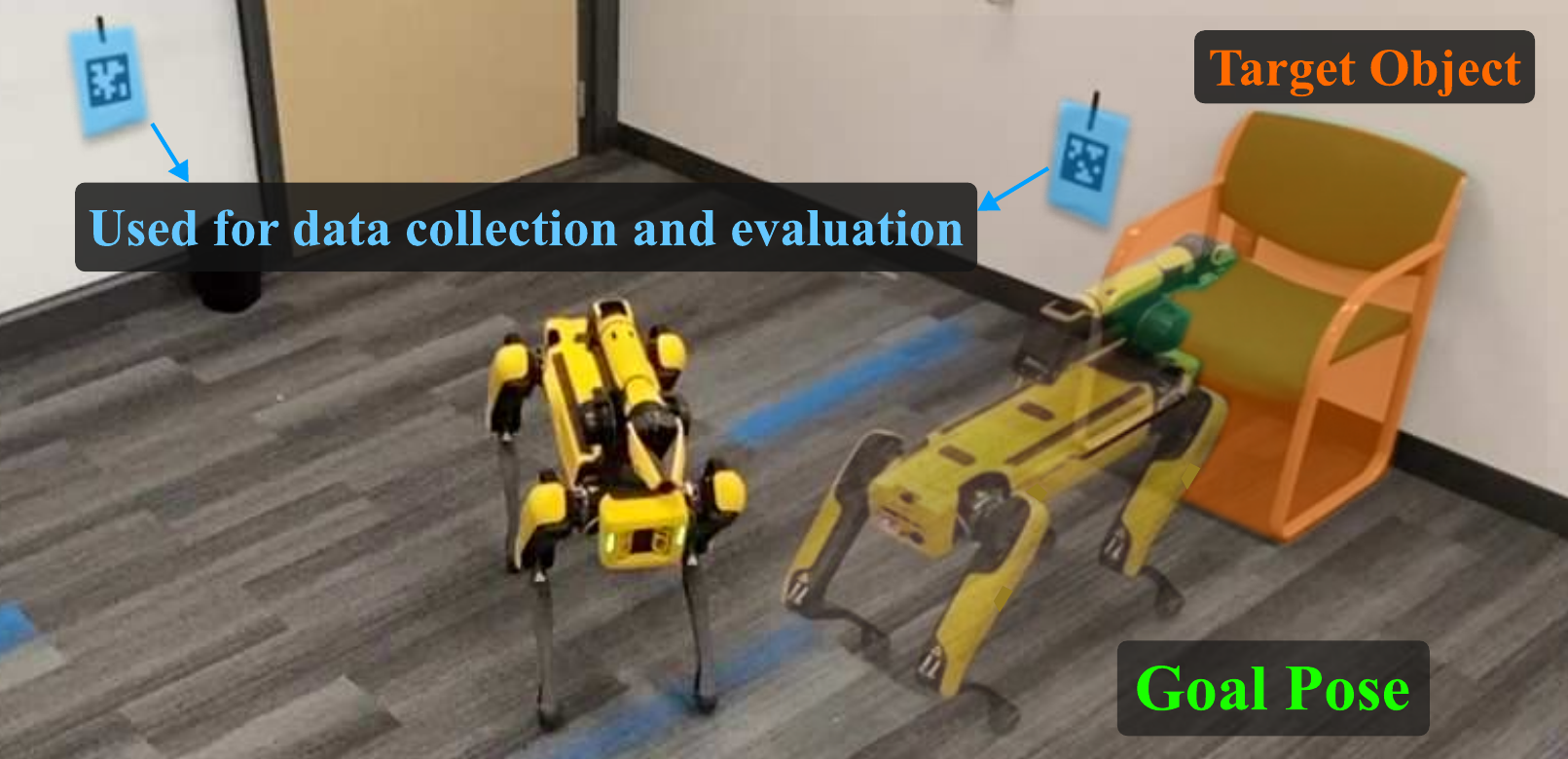}
    \caption{\footnotesize{\textbf{Training environment.} The green chair serves as the target object for training, and the goal pose of the robot indicated in the scene. AprilTags placed in the environment provide ground truth pose for automating expert demonstrations and for quantitative evaluation.}}
    \label{fig:trainging_env}
    \vspace{-0.5cm}
\end{figure}

We collected training data in an indoor environment (Fig.~\ref{fig:trainging_env}) designed to resemble everyday human living and working spaces, containing common objects such as chairs, tables, and shelves. A green chair served as the sole target object for the training dataset. Data collection was conducted using a Boston Dynamics Spot robot equipped with five onboard RGB cameras: two front-facing, one left-facing, one right-facing, and one rear-facing. The two front-facing camera images were stitched together to form a unified front view. To ensure consistent expert demonstrations, we leveraged Spot’s built in localization system, which uses AprilTag landmarks in the environment to estimate the ground truth robot pose. This localization system was used to automatically generate expert trajectories.

\subsubsection{\textbf{Automated Expert Trajectories Collection}}
We first defined the goal pose of the robot base with respect to the target object. Each trajectory was then generated by recording synchronized RGB observations($O_t$) from all cameras at every timestep, along with the corresponding discrete actions($A_t$) executed by the robot. Each action($A_t$) was computed from the difference between the robot's current pose and the defined goal position and orientation. A tolerance threshold of $0.2$ meters in translation and $\pm6\degree$ in orientation was set for determining the termination of the trajectory. These thresholds reflect Spot robot’s actuation limits, as the robot cannot reliably maintain finer positional or angular precision.

\subsubsection{\textbf{Dataset Formation and Augmentation}}
In total, we collected 715 trajectories, with trajectory length varying based on the starting pose. The final training dataset was constructed by sampling from these trajectories. Each data sample consists of a current observation($O_t$), the corresponding action($A_t$), and a pseudo-goal observation($\tilde{O}_{\text{goal},t}$). The pseudo-goal observation is taken from any future timestep in the same trajectory:

$$\tilde{O}_{\text{goal}, t} \in \{O_{t'} \mid t' \in t+1, t+2, ..., T\}$$

This sampling strategy expanded the dataset by capturing a variety of intermediate goal relationships.

\subsubsection{\textbf{Training Starting Pose Definition}}
We systematically parameterized the robot’s starting poses around the target object using three key variables: \textit{Radial Distance}, \textit{Approach Angle}, and \textit{Starting Orientation} (Fig.~\ref{fig:training_and_rollout_points}). 
\begin{itemize}[leftmargin=*]
    \item Radial Distance: The robot began at distances $\{0.3,0.45,0.6,0.9,1.2\}$ meters from the object, spanning close to far initiation points.
    \item Approach Angle: The placement ranged from $-90\degree$ to $90\degree$ in $15\degree$ increments, determining the side of the object from which the robot approached.
    \item Starting Orientation: The robot’s initial heading was varied from $-150\degree$ to $150\degree$ in $30\degree$ increments, to ensure comprehensive evaluation of initial heading variations. 
\end{itemize}

\subsection{Rollout}
\begin{figure}
  \centering
  \includegraphics[width=\linewidth]{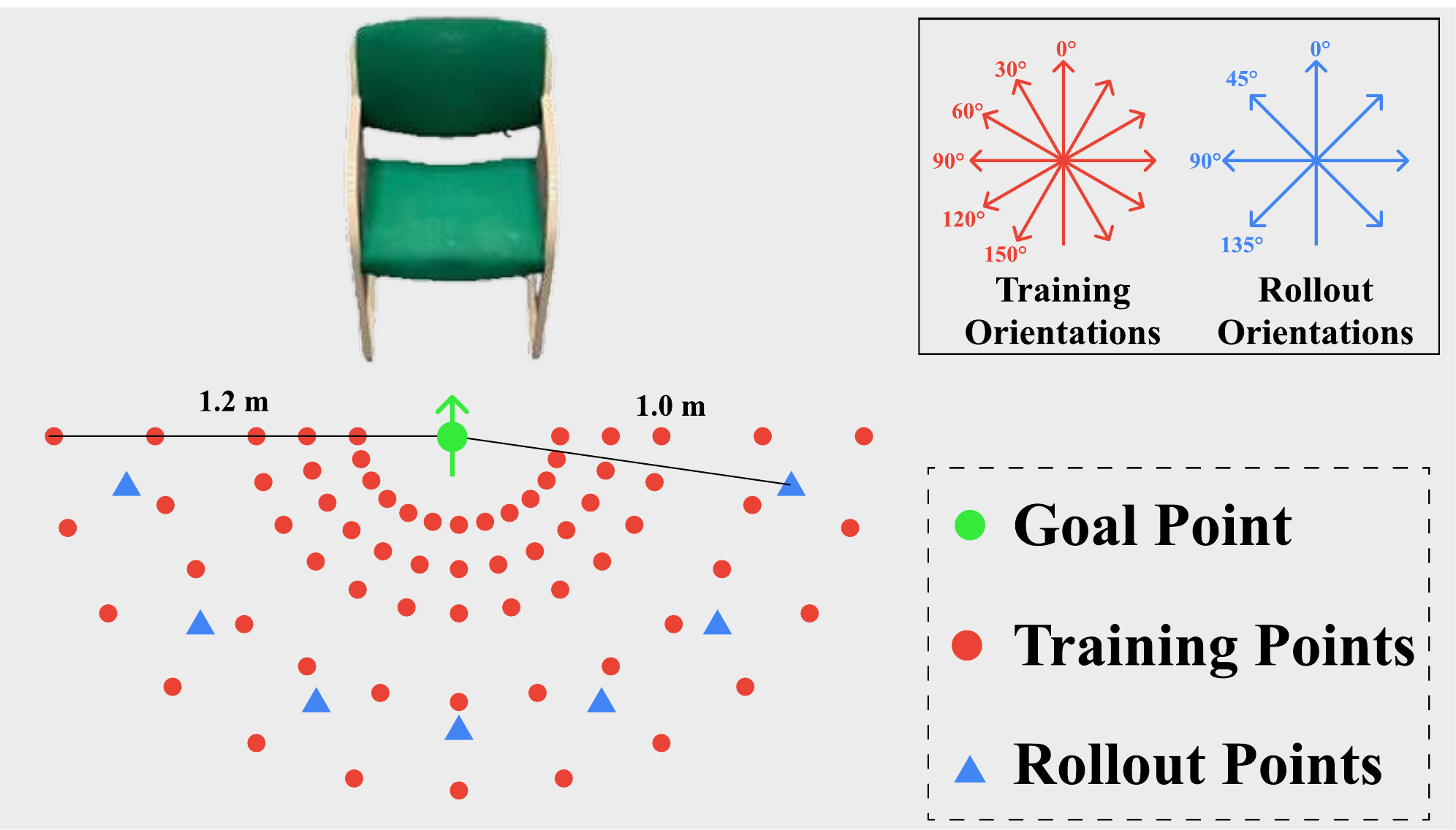}
  \caption{\footnotesize{\textbf{Starting pose distributions for training and rollout.} From each starting pose, the robot moves toward the goal pose. Training points (red) span several Radial Distances up to a maximum radius of 1.2 meters, with multiple initial orientations at each point. Rollout points (blue) are sampled at a fixed radius of 1.0 meter with their own orientation set.}}
  \label{fig:training_and_rollout_points}
  \vspace{-0.5cm}
\end{figure}

To enable objective and consistent quantitative evaluation, we used Spot robot’s built-in localization system to perform fully automated rollout experiments. We defined a set of starting poses using the same parameterization scheme as in data collection, but assigned different values to the parameters. For rollouts, the Radial Distance was fixed at 1 meter; the Approach Angles were set to [$80\degree$ $50\degree$ $25\degree$ $0\degree$ $-25\degree$ $-50\degree$ $-80\degree$]; and the Starting Orientations were set to [$135\degree$ $90\degree$ $45\degree$ $0\degree$ $-45\degree$ $-90\degree$ $-135\degree$](Fig.~\ref{fig:training_and_rollout_points}). The robot automatically navigated to each starting pose and then executed the learned policy to move toward the goal pose. A maximum duration of 100 seconds was enforced; otherwise rollouts terminated as described in Section~\ref{sec:stopping_mechanism}. Any rollout that exceeded this limit was externally terminated and marked as a failure, since successful last-meter navigation within our defined setup should complete well within this timeframe.

\subsection{Evaluation}
Since last meter navigation demands precision in both position and orientation, our evaluation explicitly measures translation and orientation error.

\subsubsection{Translation}
Translation error is measured using Spot robot’s onboard localization system. For each rollout, we compute the Euclidean distance between the robot’s final achieved position and the annotated goal position. A rollout is considered successful in translation if the final position lies within 0.3 meters of the goal location. To account for real world execution uncertainty, this threshold is slightly relaxed relative to the 0.2 meter tolerance used during dataset collection.

\subsubsection{Orientation}
Different downstream manipulation tasks impose different requirements on the robot’s final orientation. Consequently, we evaluate orientation using two complementary criteria: a ground truth criterion and a visual object facing criterion, each corresponding to a distinct manipulation need.

\begin{figure}
    \centering
    \includegraphics[width=\linewidth]{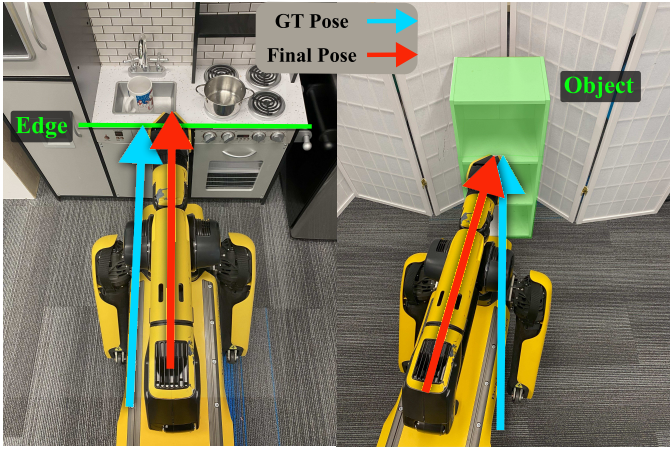}
    \caption{\footnotesize{\textbf{Examples of the two evaluation settings}: Edge Alignment (left) and Object Alignment (right). The blue arrow indicates the ground-truth (GT) pose, and the red arrow shows the robot’s final pose.}}
    \label{fig:evalutaion_explain}
    \vspace{-0.5cm}
\end{figure}

We formalize these requirements into two specific task settings:

\begin{itemize}[leftmargin=*]
    \item \textbf{Edge Alignment:} This setting utilizes the ground truth criterion. It addresses tasks where the robot must operate parallel to the linear boundary of a functional workstation, such as the front face of a kitchen sink or stove (Fig.~\ref{fig:evalutaion_explain}, left). Success is defined by how closely the final orientation matches the annotated ground truth pose. A rollout is considered successful in this edge alignment setting if it satisfies the translation success condition (i.e., 0.3 m) and its final orientation lies within a $\pm8\degree$ deviation from the ground truth.

    \item \textbf{Object Alignment:} This metric evaluates the visual object facing criterion, which is essential for tasks requiring the robot to orient itself directly toward a target, such as retrieving items from a shelf (Fig.~\ref{fig:evalutaion_explain}, right). The CoM $\mathbf{c}$ of the target mask in the final observation ($O_T$) is computed as $\mathbf{c} = \frac{1}{N} \sum_{(u,v) \in \Omega} \mathcal{M}(u,v) \cdot [u,v]^\top$, where $\mathcal{M} \in \{0,1\}^{H \times W}$ is the target's binary segmentation mask, $\Omega \subset \mathbb{Z}^2$ is the set of all pixel coordinates in the image domain, and $N$ is the total number of pixels belonging to the target mask. The alignment error is $d_{CoM} = \|\mathbf{c}_{goal} - \mathbf{c}_{final}\|_2$. A rollout is considered successful in object-alignment if it satisfies the translation condition and $d_{CoM} \leq \tau_{CoM}$, where $\tau_{CoM}$ is the maximum $d_{CoM}$ computed across the final observations of all training trajectories. The rationale is that the policy is trained via behavioral cloning to imitate the terminal states observed in the demonstrations, so we cannot expect it to achieve finer CoM precision at deployment than the training data itself exhibits at trajectory termination.
\end{itemize}

\section{Experiments}

\subsection{Baseline}
We compare our method against two baselines that we designed to represent standard approaches for vision-language policy learning, each isolating a specific architectural choice in our framework.

The first baseline, DinoTxtAttention, follows a canonical multimodal attention pipeline. In this base line, DINOv2 is used as the vision encoder to extract embeddings from both the current and goal observations. A pretrained text encoder~\cite{jose2024dinov2meetstextunified} maps the language prompt to a text embedding that matches the dimensionality of the DINOv2 embeddings. The text embedding then attends to both the goal and current visual embeddings. The resulting attended goal and current embeddings are passed through a cross-attention layer and a feedforward action head to predict the discrete actions.

The second baseline, DinoAttention, evaluates the efficacy of our explicit Score Matrix representation. This model retains the identical upstream segmentation pipeline to extract masked features for both the current and goal observations. However, instead of computing a static score matrix, it employs a standard multi-head cross-attention mechanism to model the relationship between the two views. In this setup, the masked current embedding attends to the masked goal embedding, and the resulting context-aware features are passed directly to the feed forward layer to predict the action.

\subsection{\textit{Which policy performs best for last-meter navigation?}}
\label{sec:policy_comparison}

\begin{figure*}
    \centering
    \includegraphics[width=\textwidth]{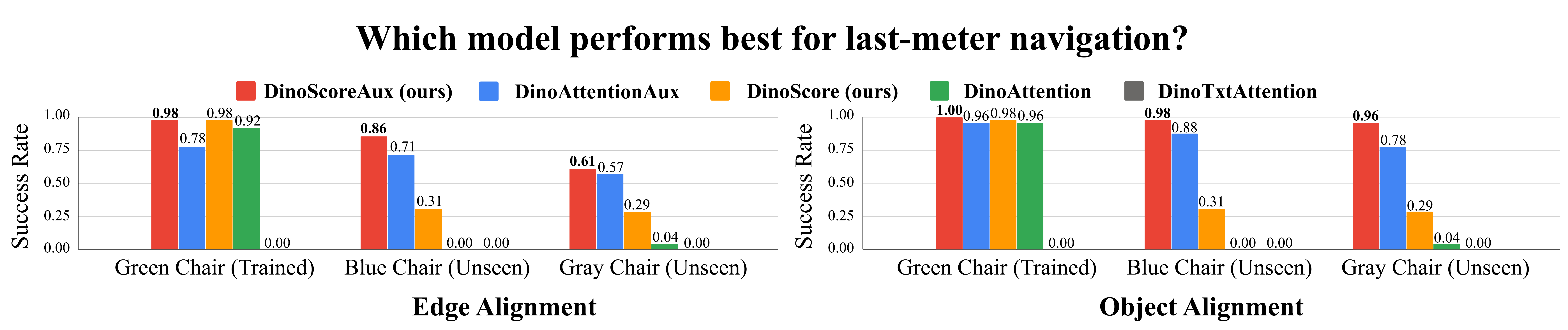}
    \caption{\footnotesize{\textbf{Success rates of different policies on last-meter navigation across seen and unseen objects.} The \textit{left} reports performance under the edge alignment setting, which evaluates success based on ground truth translation and orientation. The \textit{right} reports performance under the object alignment setting, which uses ground truth translation and the center of mass of the target mask in the final observation. Across all evaluation settings, our system \textbf{DinoScoreAux} achieves the highest success rate among all tested methods, demonstrating strong generalization from the trained green chair to unseen chair instances.}}
    \label{fig:part1}
    \vspace{-0.5cm}
\end{figure*}

We denote our proposed architecture from Section~\ref{sec:architecture} as \textbf{DinoScore}. Our comparison across policies reveals three major observations, as shown in Fig.~\ref{fig:part1}.

First, \textbf{implicit visual grounding is insufficient for this task.} The DinoTxtAttention baseline fails completely, achieving zero success on both seen and unseen objects. This failure indicates that without an explicit segmentation module, a pure vision-language attention approach cannot reliably localize the target object or disentangle it from the background when training data is limited to a single instance. Explicit object grounding is therefore a prerequisite for stable last-meter navigation in this low-data regime.

Second, \textbf{the explicit Score Matrix is critical for category-level generalization.} DinoScore achieves consistently higher success rates than DinoAttention on both the trained green chair and unseen chair instances. This performance gap highlights that the Score Matrix provides a stronger representation of the spatial relationship between the robot’s current pose and the target object. By explicitly modeling the spatial correlation between the current and goal views, the Score Matrix enables the policy to learn the geometric relationship from a single object instance and successfully generalize that understanding to the entire category of unseen objects. In contrast, the attention-based baseline struggles to capture these precise spatial relations, leading to lower success rates on unseen instances.

Third, \textbf{purely learned policies struggle with precise termination.} We observe that neither DinoAttention nor DinoScore reliably produces the stop action consecutively, even when the robot has effectively reached a manipulation-ready pose. This behavior stems from noise in the demonstration data, arising from the Spot robot's inability to execute fine positional or angular adjustments, which leads to unavoidable variation in the final states. To address this issue, we apply the auxiliary stopping mechanism described in Section~\ref{sec:stopping_mechanism}. We denote the resulting systems as DinoAttentionAux and DinoScoreAux. As shown in Fig.~\ref{fig:part1}, this mechanism is essential for converting the policy's navigational success into a successful task completion.

\subsection{\textit{How does the best policy perform in unseen environments?}}

\begin{figure*}
    \centering
    \includegraphics[width=\textwidth]{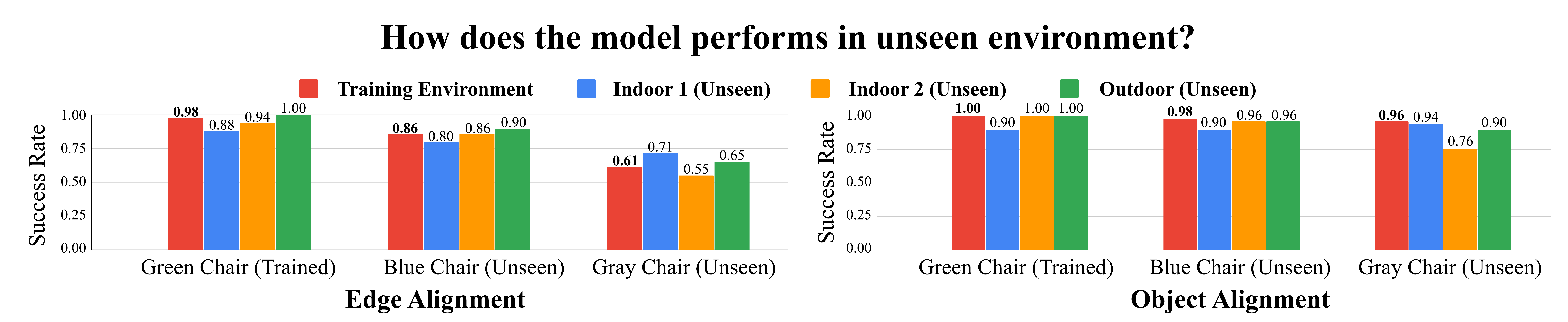}
    \caption{\footnotesize{\textbf{Success rates of our best system, DinoScoreAux, across the training environment and three unseen environments.} The \textit{left} reports performance under the edge alignment setting, which evaluates success based on ground truth translation and orientation. The \textit{right} reports performance under the object alignment setting, which uses ground truth translation and the center of mass of the target mask in the final observation. \textbf{DinoScoreAux} maintains high success rates across all environments.}}
    \label{fig:part2}
\end{figure*}

\begin{figure}
    \centering
    \includegraphics[width=\columnwidth]{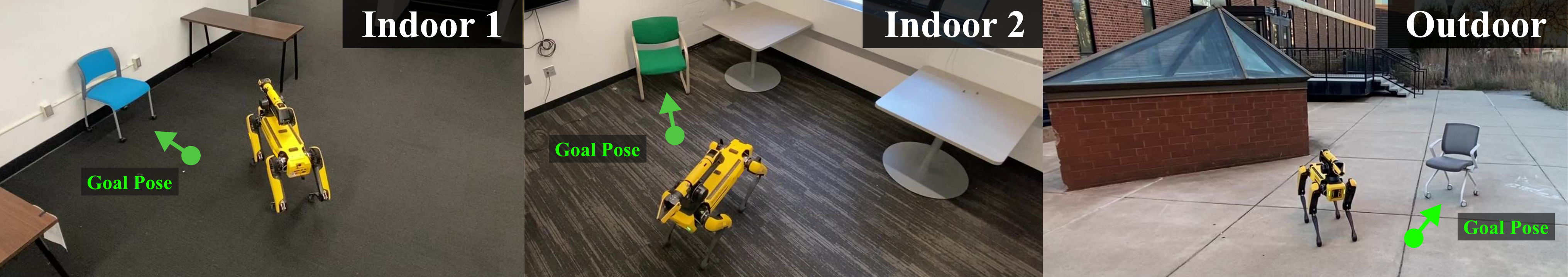}
    \caption{\footnotesize{\textbf{Unseen indoor and outdoor environments.} We evaluate the system’s generalization ability in two unseen indoor environments and one outdoor environment. In each scene, the goal pose is marked by the green point and arrow on the floor.}}
    \label{fig:env_maps}
    \vspace{-0.5cm}
\end{figure}

To assess the robustness of our best-performing policy, we evaluate its generalization across three unseen environments: two unseen indoor environments and one outdoor environment (Fig.~\ref{fig:env_maps}). This evaluation examines how well the system maintains precise last-meter navigation under varied lighting conditions, backgrounds, and scene layouts.

First, \textbf{we observe that environmental lighting is the dominant factor influencing performance.} As shown in Fig.~\ref{fig:part2}, the outdoor environment yields the highest success rates, achieving 85\% in edge alignment and 95\% in object alignment. In contrast, the indoor environments achieve lower averages of 79\% and 91\%, respectively. We attribute this discrepancy to the Spot robot's cameras, which capture higher-quality images in natural outdoor lighting. Higher image fidelity enhances the perception stack, allowing the DINOv2 encoder to extract robust features and the segmentation module to generate precise object masks. Consistent with this hypothesis, more than half of all failure trajectories in indoor scenes can be traced to incorrect or unstable segmentation masks.

Second, \textbf{qualitative analysis reveals that even in failed trajectories, the robot generally navigates in the correct direction toward the target.} This indicates that the policy successfully generalizes the global approach behavior. Failures are typically driven by two factors. First, as discussed in the previous section, the system struggles to execute the final stop action due to variance in the final states of the demonstration data, which often causes the robot to drift past the goal pose. Second, temporal inconsistency in the segmentation module prevents the decoder from receiving stable geometric cues, as the predicted masks often flicker or shift between frames.

\subsection{\textit{How does the best policy perform on unseen instances in their own environment?}}

\begin{figure*}
    \centering
    \includegraphics[width=1.0\textwidth]{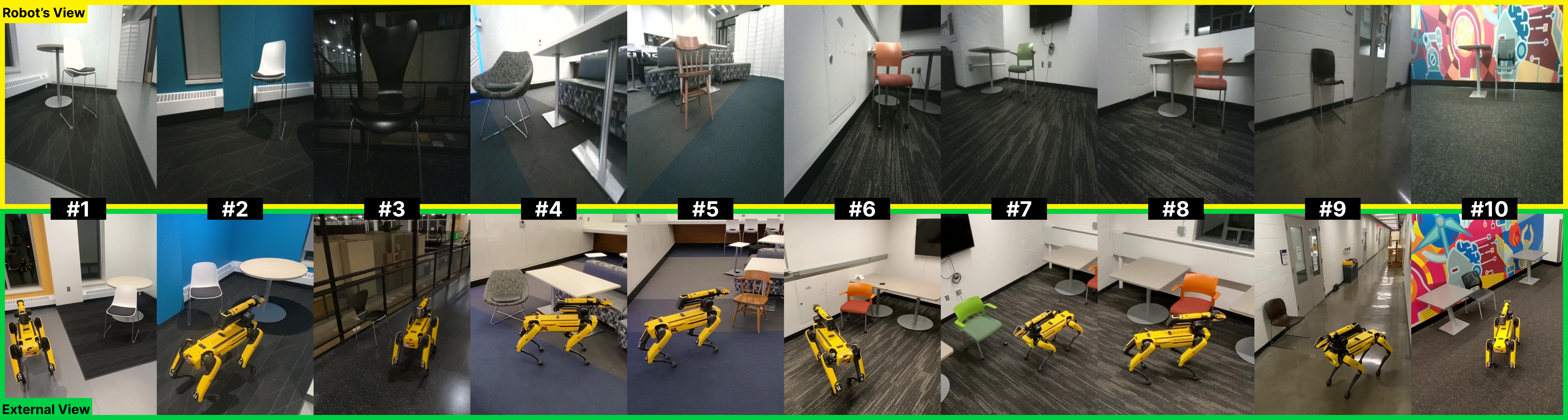}
    \caption{\footnotesize{\textbf{Ten unseen scenarios used for evaluating generalization.}}}
    \label{fig:unseen_chairs}
    \vspace{-0.5cm}
\end{figure*}

\begin{figure}
    \centering
    \includegraphics[width=\linewidth]{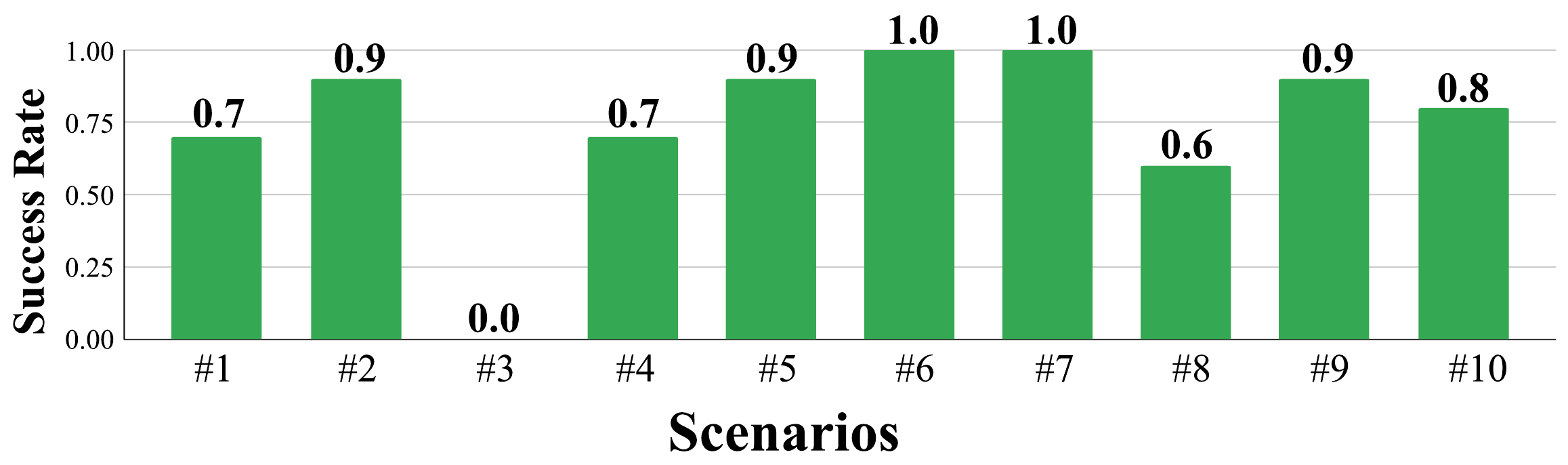}
    \caption{\footnotesize{\textbf{Instance-level generalization across unseen scenarios.} Our system achieves an average success rate of \textbf{75\%}, indicating strong transfer of its object-centric navigation strategy to novel instances of the same category.}}
    \label{fig:part3}
    \vspace{-0.5cm}
\end{figure}

To evaluate category-level generalization, we conducted an extensive study across ten distinct scenarios (Fig.~\ref{fig:unseen_chairs}). Each scenario featured a unique, unseen chair instance placed in a different location. We performed 10 rollout trials per scenario, with success rates determined by expert human verification of the final pose. As reported in Fig.~\ref{fig:part3}, the policy achieves an average success rate of 75\%, demonstrating that the learned object-centric scoring mechanism reliably transfers to novel instances of the same category without fine-tuning.

As in earlier evaluations, even the unsuccessful trajectories generally move toward the correct object, indicating that the control policy remains effective. The primary failure cases arise from inaccurate or unstable segmentation, often caused by challenging visual conditions. For example, Scenario 3 is placed in a dimly lit environment, leading to a zero \% success rate. This case further illustrates the sensitivity of the segmentation module to illumination differences and reinforces the earlier observation that lighting variations significantly affect the system.

\subsection{\textit{Does the framework generalize beyond the chair category?}}
\label{sec:cross_category}

To assess whether the framework extends beyond chairs, we evaluated on two additional object categories with markedly different visual and geometric properties: drawer organizers and suitcases. For each category, we collected demonstrations on a single instance using the same automated pipeline, trained a separate DinoScoreAux policy, and evaluated it on two unseen instances of the same category without any fine-tuning. The auxiliary stopping mechanism used identical threshold values across all categories.

Across both new categories, DinoScoreAux maintained strong performance on unseen instances, achieving an average of 0.60 in edge alignment and 0.73 in object alignment for drawer organizers, and 0.91 and 0.96 for suitcases. These results, consistent with the chair experiments, indicate that the proposed framework transfers from a single training instance to unseen instances across object categories with different shapes and appearances, supporting the generality of the approach beyond a single object type.
\section{Conclusion}
In this work, we presented an object-centric imitation learning framework for last-meter navigation that achieves manipulation-ready precision of about 0.3 meters in translation and $9\degree$ in orientation using only RGB observations. Experiments on the Spot robot show strong generalization to unseen instances and new environments, demonstrating that the learned policy reliably approaches target objects and often attains the correct final pose. The system remains constrained by its dependence on segmentation quality and unobstructed target visibility, motivating future work on integrating stronger segmentation models and improving robustness to occlusions.
\section*{Acknowledgment}
The authors would like to thank Nirshal Chandra Sekar and Adam Imdieke for their assistance with the experiments, and Zachary Chavis for his advice on the visual presentation. We also acknowledge the members of the Robotics: Perception and Manipulation Lab for their support, and the Minnesota Robotics Institute for providing seed funding the experimental facilities.

\printbibliography

\end{document}